\documentclass[final]{cvpr}

\usepackage{times}
\usepackage{epsfig}
\usepackage{graphicx}
\usepackage{amsmath}
\usepackage{amssymb}
\usepackage{diagbox}
\usepackage{soul}
\usepackage{subfig}
\usepackage[dvipsnames]{xcolor}
\definecolor{amber}{rgb}{1.0, 0.49, 0.0}

\usepackage[pagebackref=true,breaklinks=true,colorlinks,bookmarks=false]{hyperref}

\def\cvprPaperID{3383} 
\def\confYear{CVPR 2021}

\begin{document}

\title{Unsupervised Part Segmentation through Disentangling\\  Appearance and Shape}

\author{Shilong Liu$^{1}$, Lei Zhang$^{2}$, Xiao Yang$^{1}$, Hang Su$^{1}$, Jun Zhu$^{1}$\thanks{Corresponding author} \\
$^{1}$ Dept. of Comp. Sci. and Tech., BNRist Center, Institute for AI, Tsinghua-Bosch Joint ML Center\\
$^{1}$ Tsinghua University, Beijing, 100084, China  \hspace{2ex} 
$^{2}$ Microsoft Corporation \\
\small{\{liusl20, yangxiao19\}@mails.tsinghua.edu.cn}, 
\small{leizhang@microsoft.com}, 
\small{\{suhangss, dcszj\}@mail.tsinghua.edu.cn} \\
}

\maketitle

\begin{abstract}
We study the problem of unsupervised discovery and segmentation of object parts, which, as an intermediate local representation, are capable of finding intrinsic object structure and providing more explainable recognition results. Recent unsupervised methods have greatly relaxed the dependency on annotated data which are costly to obtain, but still rely on additional information such as object segmentation mask or saliency map. To remove such a dependency and further improve the part segmentation performance, we develop a novel approach by disentangling the appearance and shape representations of object parts followed with reconstruction losses without using additional object mask information.
To avoid degenerated solutions, a bottleneck block is designed to squeeze and expand the appearance representation, leading to a more effective disentanglement between geometry and appearance. 
Combined with a self-supervised part classification loss and 
an improved geometry concentration constraint, we can segment more consistent parts with semantic meanings. Comprehensive experiments on a wide variety of objects such as face, bird, and PASCAL VOC objects demonstrate the effectiveness of the proposed method.

\end{abstract}

\begin{figure}[t]
\begin{center}
  \includegraphics[width=1.0\linewidth]{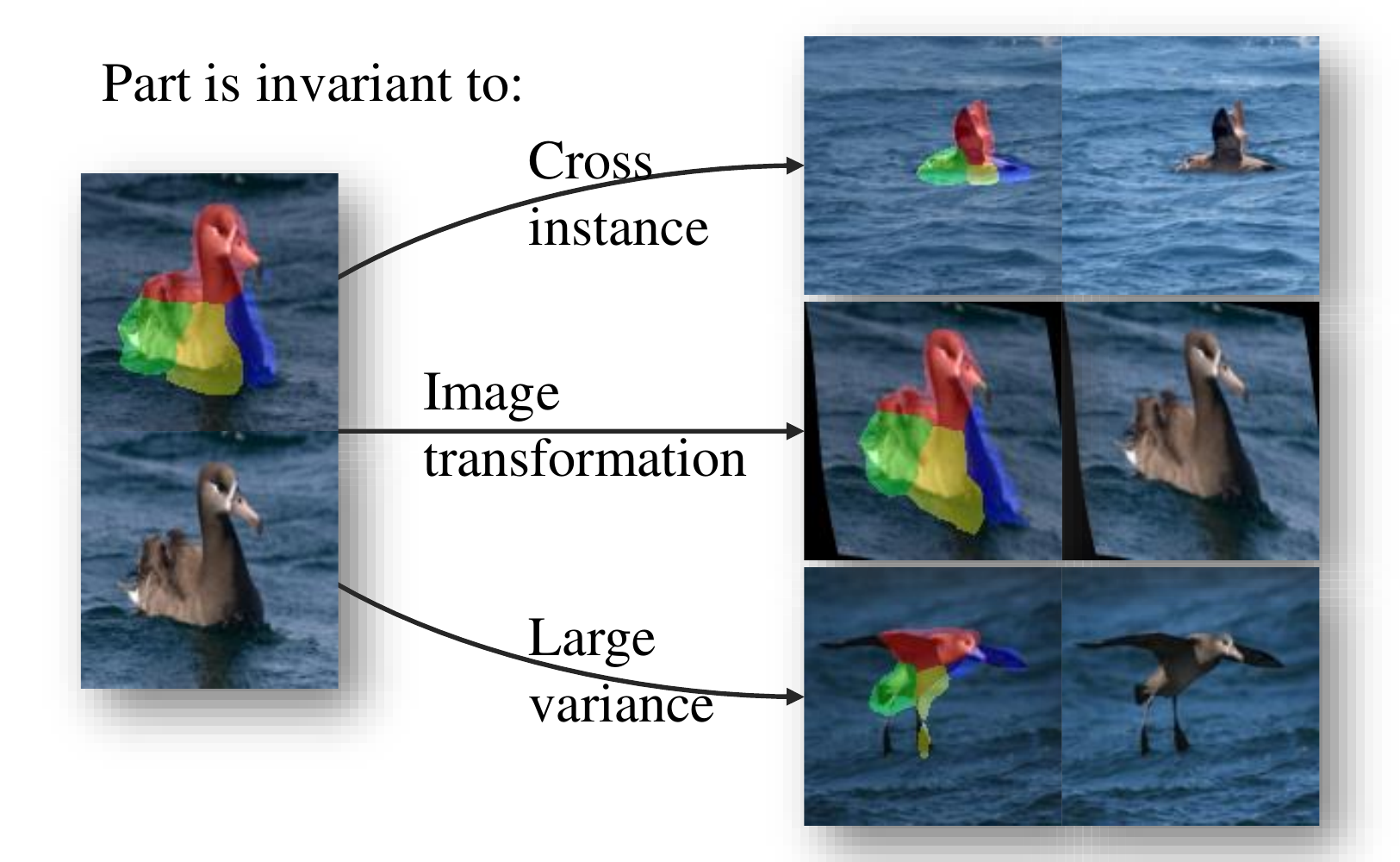}
\end{center}
\vspace{-4ex}
  \caption{Semantic consistency of object parts. Segmentation regions of the same part should be semantically consistent across object instances and robust to appearance and shape changes. Best view in colors.
  }
\vspace{-3ex}
\label{fig:consistency}
\end{figure}

\section{Introduction}
\label{sec:1}

Object parts and landmarks are two widely used intermediate representations of object local structures, which have received increasing attention recently for their robustness to the variations of viewpoint and appearance~\cite{ hung:CVPR:2019}. While there have been many related works on object landmark detection by either supervised learning~\cite{zhang2014facial, ranjan2019hyperface}
or unsupervised discovery~\cite{jakab2018unsupervised, zhang2018unsupervised}
, semantic part segmentation is relatively less studied~\cite{hung:CVPR:2019} due to the costly efforts required for data annotation. 

This paper primarily focuses on unsupervised semantic part segmentation, as object parts provide richer information about intrinsic object structures which are complementary to landmark points. To alleviate the efforts in annotations, we investigate the unsupervised learning strategy for automatic semantic part segmentation~\cite{ hung:CVPR:2019}. It has also shown an encouraging impact on single-view 3D object reconstruction by using self-supervised learned semantic parts which enforces semantic consistency between the reconstructed 3D mesh and original 2D images~\cite{li2020self}.

A good object part segmentation should satisfy two essential constraints on \textit{semantics} and \textit{geometry}, which correspond to object appearance and shape, respectively. The \textit{semantic constraint} means that segmentation regions of the same part should be semantically consistent across object instances and robust to appearance and shape changes. The \textit{geometric constraint} means a part segmentation region should be locally connected and the union of all parts should entirely cover the corresponding object. These two constraints can be seen in Fig. \ref{fig:consistency}, in which the parts with the same semantic information (e.g., heads of bird) are covered by the same segmentation color (e.g., red) in different images, and the union of all parts provides a good coverage over an object. Hung \etal~\cite{hung:CVPR:2019} also discuss similar constraints including geometric concentration, robustness to variations, semantic consistency, and objects as union of parts, and accordingly develop an unsupervised solution for semantic part segmentation. However, for the objects-as-union-of-parts constraint, their solution lacks a principled way to enforce such a constraint and has to take a dependency on saliency map, which is not always reliable and has to be replaced with object segmentation mask in some cases~\footnote{For example, for CUB-Birds, the open-source implementation of SCOPS uses ground truth silhouettes rather than saliency maps and crops birds w.r.t bounding boxes rather than using the original images. See \url{https://github.com/NVlabs/SCOPS\#scops-on-caltech-ucsd-birds}}.
This dependency makes it hard to scale up with more data or to other object categories.
\label{sec:analysis_method}

To relax the dependency on saliency map (or object segmentation mask) and further improve the performance of part segmentation, we develop a novel unsupervised learning approach by disentangling the appearance and shape representations of semantic parts, followed with reconstruction losses without using any additional object segmentation mask or saliency map \footnote{Note that we still assume the availability of object bounding boxes, which are used in all unsupervised object landmark/part learning works to bound the learning problem to single object images.}. The idea of disentanglement and synthesis is inspired by \cite{jakab2018unsupervised}, despite that it was proposed for automatic landmark point discovery whereas this work is for semantic part segmentation. By disentangling the appearance and shape representations of a target object category, we can effectively add semantic and geometric constraints on the appearance part and the shape part, respectively. 

More specifically, our framework is an encoder-decoder based deep neural network. The network firstly takes a pair of images which only differ by geometric transformations, and then uses two encoders to extract the appearance and shape representations of the two images. After exchanging the appearance representations between two images, a decoder is used to reconstruct the input images to close the loop for unsupervised learning.
To avoid degenerated solutions, we design a bottleneck block to squeeze and expand the appearance representation, leading to a more complete disentanglement between appearance and shape, which in return yields a better shape representation for part segmentation. On the squeezed appearance representation, we apply a classification loss to learn distinctive part features while enforcing cross-instance semantic consistency. On the disentangled shape representation, we use an improved geometric concentration loss with a special treatment for background pixels. This concentration loss, together with the reconstruction loss, essentially ensures the geometric constraint. To validate the effectiveness of the proposed framework, we perform comprehensive experiments on a wide variety of objects such as face, bird, and PASCAL VOC objects, and obtain superior part segmentation results compared to previous works.

Our contribution can be summarized as follows:
\begin{itemize}
\setlength{\itemsep}{0pt}
\setlength{\parsep}{0pt}
\setlength{\parskip}{0pt}
    \item  To our best knowledge, this is the first attempt that can unsupervised learn semantic part segmentation without using any addition object segmentation mask (or saliency map).
    \item By utilizing a squeeze-and-expand bottleneck block, the proposed method can effectively disentangle the appearance and shape representations of objects and remove the dependency of using additional segmentation masks.
\end{itemize}

\section{Related Work}

\subsection{Unsupervised Landmark Discovery}
\label{sec:2.1}

There have been many efforts to learn landmarks in a supervised manner~\cite{zhu2015face, zhang2016learning, yu2016deep, xiao2017recurrent, ufer2017deep, chen2019deep, ranjan2019hyperface, wang2019joint}. However, supervised landmark detection is largely limited by the availability of annotated data and mainly works for popular categories, such as face~\cite{liu2015faceattributes, koestinger11a}, bird~\cite{WahCUB_200_2011}, \textit{etc}. To avoid relying massive annotated data, unsupervised landmark discovery has attracted much more attention in recent years. Thewlis \etal~\cite{thewlis2017unsupervised_iccv} proposed to address landmark discovery by using the {equivariance} constraint and then expanding the formulation to a dense situation~\cite{thewlis2017unsupervised}. 
Jakab \etal~\cite{jakab2018unsupervised} proposed to learn landmarks by first disentangling pose and appearance of an image and then recombining them for synthesizing a new image. We term this approach as a \textit{disentanglement and synthesis paradigm}. This work shows that combining disentanglement with synthesis can effectively lead to the desired equivariance property, which has inspired many follow-up works, including this paper. Many subsequent works suggest improvement upon ~\cite{jakab2018unsupervised}. For example, Thewlis \etal~\cite{Thewlis2019a} proposed to add a vector exchanging procedure for better semantic consistency across instances. Jakab \etal~\cite{jakab2020self} then tried to learn human-interpretable landmarks from unpaired labels and Xu \etal~\cite{xu2020unsupervised} designed a cycle framework to learn from unpaired data. Nevertheless, such works are limited to landmark discovery, whereas our work is to obtain part segmentation from unlabeled data, which is a more challenging problem.

\subsection{Part Segmentation}
Our work is closely related to weakly supervised object co-segmentation. Co-segmentation aims at identifying pixels of interested foreground objects from background. There have been many related works on this subject~\cite{hochbaum2009an, joulin2010discriminative, batra2010icoseg, sidi2011unsupervised, chai2011bicos, meng2012object, han2018robust}, but they mainly deal with semantic information at the object level rather than at the part level. Besides, most works need to take a whole dataset as input and perform co-segmentation, while we are more interested in learning a model that can predict part level semantic information, shown as a part segmentation map, for each image. 

Several recent works have attempted to learn part-level information in an unsupervised manner. Hung \etal~\cite{hung:CVPR:2019} proposed to learn part features that are semantically consistent across images and have achieved good results in their paper. However, it takes a dependency on object saliency map, which is not always reliable and has to be replaced with object segmentation mask in some cases, to distinguish a foreground object from background. In this work, we remove this dependency and make semantic part segmentation applicable to more object categories. Lorenz \etal~\cite{lorenz2019unsupervised} proposed to disentangle object shape and appearance to obtain a part modeling result, which has a similar goal to this work. However they make an elliptical assumption for each part and do not assign a part label for each pixel. Thus this approach does not work well for images whose main object occupies a large area, like human faces, as it cannot generate a complete segmentation map.

\begin{figure*}
\vspace{-1ex}
\begin{center}
\includegraphics[width=0.9\linewidth]{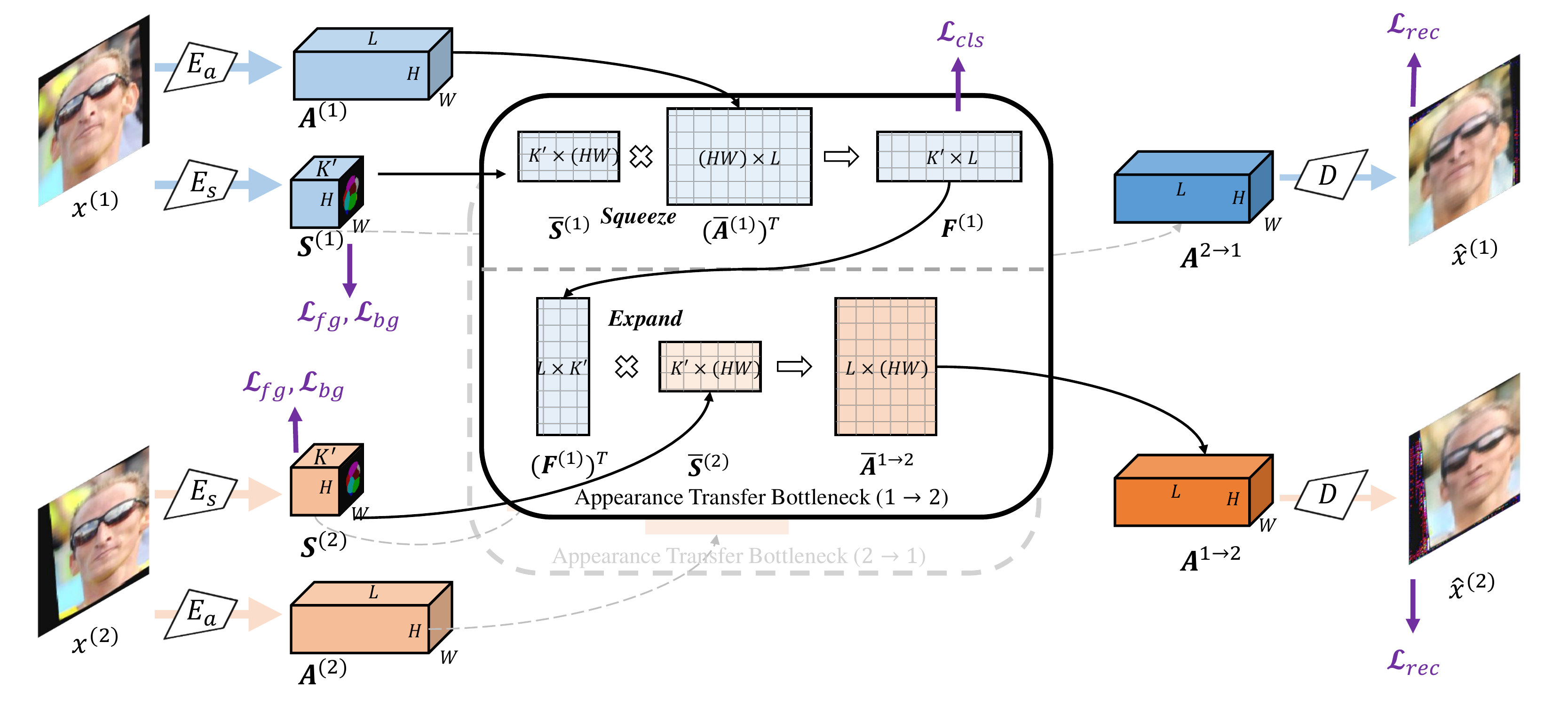}
\end{center}
\vspace{-3ex}
  \caption{The overall framework of our model.  The network firstly takes a pair of images which only differ by geometric transformations and then uses two encoders to extract the appearance $\mathbf{A}^{(i)},i=1,2$ and shape representation $\mathbf{S}^{(i)},i=1,2$ of the two images. In the middle are two bottleneck blocks, each for squeezing the appearance representation and combining it with the shape representation of another image by an expansion operation. After exchanging the appearance representations between two images, a decoder is used to reconstruct the input images to close the loop for unsupervised learning. 
  }

\label{fig:framework}
\vspace{-1ex}
\end{figure*}

\section{Methodology}
\label{sec:method}

As discussed in Sec. \ref{sec:analysis_method}, we hope the learned part segmentation results could meet both the \textit{semantic constraint} and \textit{geometry constraint} without additional saliency map or segmentation mask. To apply the two constraints on the appearance and shape features separately, we leverage an encoder-decoder neural network with a squeeze-and-expand bottleneck block in the middle to disentangle the appearance and shape representations, which will be described in Sec. \ref{sec:3.1}. To ensure the semantic constraint, we use a reconstruction loss and a classification loss, which will be detailed in Sec. \ref{sec:3.2} and Sec. \ref{sec:3.3} respectively. For the geometry constraint, we improve the concentration loss and describe the details in Sec. \ref{sec:3.4}.

\subsection{Overall Framework}
\label{sec:3.1}
The overall framework is an encoder-decoder based deep neural network, which includes three modules: a part segmentation encoder ${E}_s$, an appearance encoder ${E}_a$, and a decoder $D$ for image generation. Between the two encoders and the decoder, a bottleneck block is designed to squeeze and expand the appearance representation, for a more effective disentanglement between appearance and shape.
The whole framework is shown in Fig. \ref{fig:framework}.

To learn such a neural network, we assume an image collection of the interested object category is available as training data. Given an image from the training dataset, we firstly construct a pair of images
which only differ by geometric transformations. That is, the transformed objects included in the two images have the same appearance but different shape representations. As suggested in previous works~\cite{jakab2018unsupervised, lorenz2019unsupervised}, we use the thin-plate-spline transformation (TPS) ~\cite{duchon1977splines, wahba1990spline} to obtain two transformed images $x^{(1)},x^{(2)}$.
Using $x^{(1)}$ and $x^{(2)}$ as inputs, we can extract an appearance feature map $\mathbf{A} \in \mathbb{R}^{L\times H\times W}$ and a part segmentation map  $\mathbf{S}\in \mathbb{R}^{(K+1)\times H\times W}$ (corresponding to shape) by  $E_a$ and $E_s$ for each image as 
\begin{equation}
\label{eq:encode}
    \begin{aligned}
        \mathbf{A}^{(i)}=E_a(x^{(i)}),\ \ \ \mathbf{S}^{(i)} = E_s(x^{(i)}), \    i=1,2\\
    \end{aligned}
\end{equation}
\noindent
where $K$ denotes the number of object parts, $L$ denotes the dimension of appearance features, and $H\times W$ is the resolution of two output tensors, including both the appearance feature map and the part segmentation map.
The $0$-th channel is used for representing the background region and the $1$-st to $K$-th channels are used for representing $K$ object parts. For notational convenience, we denote $K'=K+1$.
Moreover, each part segmentation map $\mathbf{S}^{(i)},i=1,2$ is normalized along the channel dimension by $\operatorname{softmax}$.

\textbf{Appearance Transfer Bottleneck.} 
To disentangle the appearance and shape representations, reconstruction is normally performed to provide a self-supervised loss \footnote{In the paper, we mainly use ``self-supervised'' to refer to a specific loss, but use ``unsupervised'' to refer to the overall part segmentation problem which does not require supervised annotations.} after exchanging the shape (or appearance) representations between two input samples. However, directly performing reconstruction will lead to a degenerated solution. For example, the part segmentation map $\mathbf{S}$ may contain some appearance information as well as shape information, making itself capable of reconstructing the original image, which is an undesired trivial solution. Inspired by ~\cite{jakab2018unsupervised} and ~\cite{lorenz2019unsupervised}, we design a simple appearance transfer bottleneck to \textit{squeeze} and \textit{expand} the part appearance features, aiming at generating a rendered appearance feature map that contains appearance feature from only one image and shape information from another image.

In the \textit{squeeze} step (Eq.~\eqref{eq:squeeze}), each channel in the part segmentation map can be regarded as an attention map, which is used to weighted average the learned appearance features over all pixels. This step reduces the dimension of the appearance feature and squeezes out the shape information of $x^{(1)}$. In the \textit{expand} step (Eq.~\eqref{eq:expand}), the squeezed part appearance feature of $x^{(1)}$ is expanded according to the part segmentation map of $x^{(2)}$, treating each response channel as an attention map to generate a rendered appearance feature map. 
This appearance transfer bottleneck, as well as the reconstruction, effectively helps avoid degenerated disentanglement learning.

For the sake of simplicity, we only write out the process of transferring the appearance of $x^{(1)}$ to combine with the shape of $x^{(2)}$, and it is straightforward to generalize to the inverse direction. 

First, we reshape the part segmentation map $\mathbf{S}^{(1)}$ to $\Bar{\mathbf{S}}^{(1)}\in \mathbb{R}^{K'\times (HW)}$ and the appearance feature map $\mathbf{A}^{(1)}$ to $\Bar{\mathbf{A}}^{(1)}\in \mathbb{R}^{L\times(HW)}$ for notational convenience. Then, we squeeze the part appearance feature by using the part segmentation map to obtain $\mathbf{F}^{(1)}\in \mathbb{R}^{K'\times L}$ 
as follows to push out any shape information:

\begin{equation}
\label{eq:squeeze}
\begin{aligned}
\mathbf{D}&=\operatorname{diag}(\Bar{\mathbf{S}}^{(1)}\mathbf{1}_{HW}), \\
\mathbf{F}^{(1)} &= \mathbf{D}^{-1}\Bar{\mathbf{S}}^{(1)}(\Bar{\mathbf{A}}^{(1)})^T,
\end{aligned}
\end{equation}

\noindent
where $\mathbf{1}_{HW}$ is a vector of ones with dimension $HW$ and $\mathbf{D}^{-1}$ is to perform row normalization on $\Bar{\mathbf{S}}^{(1)}$. 

For the next step, we expand the squeezed part appearance feature $\mathbf{F}^{(1)}$ of $x^{(1)}$ according to the shape (part segmentation map) of the other image $x^{(2)}$. After reshaping $\mathbf{S}^{(2)}$ to $\Bar{\mathbf{S}}^{(2)}\in \mathbb{R}^{K'\times (HW)}$, we can obtain the rendered appearance feature map $\Bar{\mathbf{A}}^{1\to 2}\in \mathbb{R}^{L\times (HW)}$ as

\begin{equation}
\label{eq:expand}
\Bar{\mathbf{A}}^{1\to 2} = (\mathbf{F}^{(1)})^T \Bar{\mathbf{S}}^{(2)}.
\end{equation}

By reshaping $\Bar{\mathbf{A}}^{1\to 2}$ back to ${\mathbf{A}}^{1\to 2}\in \mathbb{R}^{L\times H\times W}$, we get the rendered appearance feature map ${\mathbf{A}}^{1\to 2}$ which contains the appearance information of image $x^{(1)}$ only and the shape information of $x^{(2)}$ alone, as we expected.
The inverse direction for ${\mathbf{A}}^{2\to 1}$ can be derived in a similar way. 

\textbf{Image Generation.}
Using the rendered appearance feature maps $\mathbf{A}^{2\to 1}$ and $\mathbf{A}^{1\to 2}$ as inputs, we can obtain two reconstructed images using the decoder $D$, as shown in the right part of Fig. \ref{fig:framework}. 

\begin{equation}
\label{eq:rx}
\hat{x}^{(1)} = D(\mathbf{A}^{2\to 1}), \ \ \ 
\hat{x}^{(2)} = D(\mathbf{A}^{1\to 2}).
\end{equation}

\subsection{Reconstruction Loss}
\label{sec:3.2}
To ensure the \textit{semantic constraint} being applied effectively on the disentangled appearance and shape representations, we use reconstruction as self-supervised signals. Some previous works ~\cite{jakab2018unsupervised, lorenz2019unsupervised} have shown that the disentangling and synthesis paradigm could help learn features with the desired semantic consistency property, which is also termed as {equivariance}.
Our reconstruction loss is defined as

\begin{equation}
\label{eq:perceptual}
\mathcal{L}_{rec} = \mathcal{P}(x^{(1)}, \hat{x}^{(1)}) + \mathcal{P}(x^{(2)}, \hat{x}^{(2)}),
\end{equation}

\noindent
where $\mathcal{P}$ represents the perceptual loss~\cite{johnson2016perceptual, chen2017photographic} as in previous works~\cite{jakab2018unsupervised, xu2020unsupervised}.
\subsection{Part Classification Loss}
\label{sec:3.3}
Similar to semantic segmentation, part segmentation can also be considered as a multi-class classification problem to assign a class label to each pixel. Although we do not have part labels for each pixel, we can perform classification on the squeezed part appearance feature $\mathbf{F}\in \mathbb{R}^{K'\times L}$.
That is, the $k$-th part feature $\mathbf{F}_k\in \mathbb{R}^{L}$ should be classified correctly to the $k$-th class. 

Such a self-supervised classification loss not only helps learn distinctive features for different parts, but also implies \textit{semantic consistency} across object instances, which leads to the desired feature invariance property. Inspired by recent progress in face recognition feature learning~\cite{wen2016a, liu2017sphereface, wang2018cosface, guo2018oneshot, deng2019arcface}, to further reduce the intra-class variance, we adopt the ArcFace loss~\cite{deng2019arcface}, which can be formulated as: 

\begin{equation}
\label{eq:arcface}
\mathcal{L}_{cls} = \frac{1}{K}\sum_{i=1}^{K}\log \frac{e^{s(\cos (\theta_{i,i}+m))}}{e^{s(\cos (\theta_{i,i}+m))}+\sum_{j=1,j\neq i}^{K}e^{s \cos \theta_{j,i}}}.
\end{equation}

The background channel is excluded from the loss for its inconsistency across instances. To implement this loss function, we add a linear classification layer in the model, which has parameters $W\in \mathbb{R}^{K\times L}$ representing $K$ feature basis of dimension $L$. In the loss function, $\theta_{j,i}$ is the angle between part feature $F_{i}\in \mathbb{R}^{L}$ and basis $W_{j}\in \mathbb{R}^{L}$, and $s$ and $m$ are two constants representing radius scaling and angular margin penalty. 

Unlike previous image-level constraints, we apply self-supervised classification constraints directly on object parts. To our best knowledge, this is the first time such a constraint is used in unsupervised keypoint discovery or part segmentation. The use of this constraint effectively leads to distinctive part features while keeping cross-instance semantic consistency, which further helps improve the part segmentation performance as demonstrated in Sec. \ref{Sec:ablation}.

\subsection{Foreground/Background Concentration Loss}
\label{sec:3.4}
As mentioned in Sec. \ref{sec:analysis_method}, the learned segmentation map should meet the \textit{geometry constraint} as well. Our method requires each pixel to be assigned with a part label. Hence the requirement of an object being entirely covered by the union of all parts is easy to implement. 

For the concentration constraint which requires each object part to be locally connected, we follow the concentration loss as proposed in \cite{hung:CVPR:2019} for foreground object pixels but make a special modification for background pixels as they are scattered around an object and do not follow the concentration constraint. For foreground pixels:
\begin{equation}
\label{eq:foreground}
\mathcal{L}_{fg}=\frac{1}{K}\sum_{k=1}^{K} \sum_{u,v} \left\lVert \begin{bmatrix}u \\ v\end{bmatrix}  - \begin{bmatrix}c_u^k \\ c_v^k\end{bmatrix} \right\rVert_2^2\cdot \mathbf{S}_{k,u,v}/ z_k,
\end{equation}
\noindent
where $c_u^k$ and $c_v^k$ are the coordinates of the $k$-th part center along axis $u,v$, and $z_k$ is a normalization term. $c_u^k$, $c_v^k$, and $z_k$ can be calculated as follows:
\begin{equation}
\label{eq:center}
c_u^k=\sum_{u,v}u\cdot \mathbf{S}_{k,u,v}/z_k,\ 
c_v^k=\sum_{u,v}v\cdot \mathbf{S}_{k,u,v}/z_k,
\end{equation}

\begin{equation}
\label{eq:norm_z}
z_k = \min\{\sum_{u,v}\mathbf{S}_{k,u,v}, 1\}.
\end{equation}

\noindent
Note that Eq.~\eqref{eq:norm_z} is slightly different from the definition of $z_k$ in ~\cite{hung:CVPR:2019}, as we found it works better for avoiding degenerated solutions in our framework.  

The foreground concentration loss works well for foreground pixels, but it is inappropriate for background pixels which are scattered around an object and lack semantic consistency. 
~\cite{hung:CVPR:2019} is not suffered from this issue because it has an extra input of saliency map to bound the region of interested object. However, as we want to remove the dependency on an extra saliency map input, we propose a background concentration loss as follows to separate out background pixels. As the channel $0$ of $\mathbf{S}$ is regraded as background, we can write the loss as:
\begin{equation}
\label{eq:background}
\mathcal{L}_{bg}= \sum_{u,v} \left\lVert h(\langle u, v\rangle) \right\rVert^2\cdot \mathbf{S}_{0,u,v}/ z_0,
\end{equation}
\noindent
where $h(\langle u,v \rangle)$ is the distance between the pixel $\langle u,v \rangle$ and its nearest boundary which can be calculated as
\begin{equation}
\label{eq:dist_boundry}
h(\langle u,v \rangle) = \min \{u, W-u, v, H-v\}.
\end{equation}

Compared with the foreground concentration loss as in Eq.~\eqref{eq:foreground}, this constraint encourages background pixels to be close to image boundaries. 
This constraint is also based on the assumption that the target object has been cropped and roughly centered with the help of an object bounding box, as mentioned in Sec. \ref{sec:3.1}, which is a commonly used assumption in unsupervised object part learning. Since it is hard to apply a semantic consistency constraint for diverse background regions, this geometric constraint helps the proposed algorithm focus on the main object in the center of an image.

\subsection{Final Loss Function}
The overall training loss is to minimize the weighted combination of all the losses in Eq.~\eqref{eq:perceptual},~\eqref{eq:arcface},~\eqref{eq:foreground}, and~\eqref{eq:background}:
\begin{equation}
\label{eq:overall_loss}
\mathcal{L}_{sum} = \lambda_{rec}\mathcal{L}_{rec} + \lambda_{cls}\mathcal{L}_{cls} +  
 \lambda_{fg}\mathcal{L}_{fg} + \lambda_{bg}\mathcal{L}_{bg},
\end{equation}

\noindent
where $ \lambda_{rec}, \lambda_{cls}, \lambda_{fg}$ and  $\lambda_{bg}$ are hyper parameters for balancing the final loss. 

\section{Experiments}
\label{sec:exp}

To evaluate the proposed approach, we conduct experiments on several datasets, including CelebA~\cite{liu2015faceattributes} and AFLW~\cite{koestinger11a} for human faces, CUB~\cite{everingham2010the} for birds, and PASCAL VOC~\cite{wah2011the} for other common object categories. As there is no ground truth for unsupervised part segmentation, we mainly follow recent works~\cite{thewlis2017unsupervised_iccv, zhang2018unsupervised, hung:CVPR:2019} and use landmark regression as a proxy metric to evaluate the performance of part segmentation for datasets with landmark annotations, e.g. CelebA, AFLW, and CUB. For PASCAL VOC which does not have landmark annotations, we evaluate the aggregated part segmentations with the foreground segmentation using the IOU (intersection over union) metric.

The experiments on human faces (including CelebA and AFLW) and common objects (including CUB and VOC) are detailed in Sec. \ref{sec:exp_face} and Sec. \ref{sec:exp_cub_voc}, respectively. We also present qualitative results of appearance and shape disentanglement  and transfer on DeepFashion~\cite{liuLQWTcvpr16DeepFashion} in Sec. \ref{sec:exp_fashion}.

\subsection{Implementation Details}
\label{sec:imp_details}
As mentioned in Sec \ref{sec:method}, our framework consists of three modules: a part segmentation encoder $E_p$, an appearance encoder $E_a$, and a decoder $D$, all of which are implemented as deep convolutional networks using PyTorch.

We use all layers before $\operatorname{layer4}$ (as denoted in the PyTorch implementation) of an unpretrained ResNet34 as the appearance encoder, except that we set the stride of $\operatorname{layer3}$ to $1$. For the part segmentation encoder, we replace the backbone model with ResNet18 (using all layers before $\operatorname{layer4}$ and setting the stride of $\operatorname{layer3}$ to $1$ as well) and add an extra $1\times 1$ convolution layer at the end of the backbone.
We resize all the images to $128\times 128$ as input resolution, and the size of the output tensor is $H\times W=32\times 32$. For different tasks, we will use different $K$. 
Our generator is composed of seven convolution blocks, each containing a convolution or transposed convolution layer, a batch normalization layer and a ReLU layer, and an extra $3\times 3$ convolution layer at the end of the generator.
Of these seven blocks, the second and fourth contain $4\times 4$ transposed convolution layers for upsampling, and the rest contains $3\times 3$ convolution layers.

We initialize all model weights with random Gaussion noise ($\operatorname{sigma}=0.01$), and use Adam~\cite{kingma2014adam} as our optimizer, with $(\beta_1, \beta_2)=(0.9, 0.999)$, a weight decay of $5\times 10^{-4}$, and a learning rate of $1\times 10^{-4}$. 
For the ArcFace loss, we set the constants $s$ to 20 and $m$ to 0.5. More implementation details are available in the supplementary materials.

\subsection{Evaluation on Human Faces}
\label{sec:exp_face}
We first test our method on two commonly used human face dataset: CelebA~\cite{liu2015faceattributes} and AFLW~\cite{koestinger11a}. CelebA is a large-scale face dataset containing more than 200K images of celebrities collected from Internet. Each image in CelebA has 40 binary annotations, a bounding box and five landmark coordinates for each face. Our setting is the same as that in ~\cite{hung:CVPR:2019}. We use the \textit{wild image} for training and testing, and filter out the images with face covering less than $30\%$ of the pixel area. Since unsupervised results cannot be measured directly, most previous works~\cite{thewlis2017unsupervised,jakab2018unsupervised,lorenz2019unsupervised} use proxy tasks for evaluation, such as landmark regression. Following \cite{hung:CVPR:2019}, we first convert part segmentations into landmarks by taking part centers as in Eq.~\eqref{eq:center} and then learn a linear regressor to map the converted landmarks to ground truth landmarks and evaluate the regression error on test data. The result can be found in Table \ref{tb:celeba}. 

\begin{figure}[t]
\vspace{-1ex}
\begin{center}
  \includegraphics[width=0.9\linewidth]{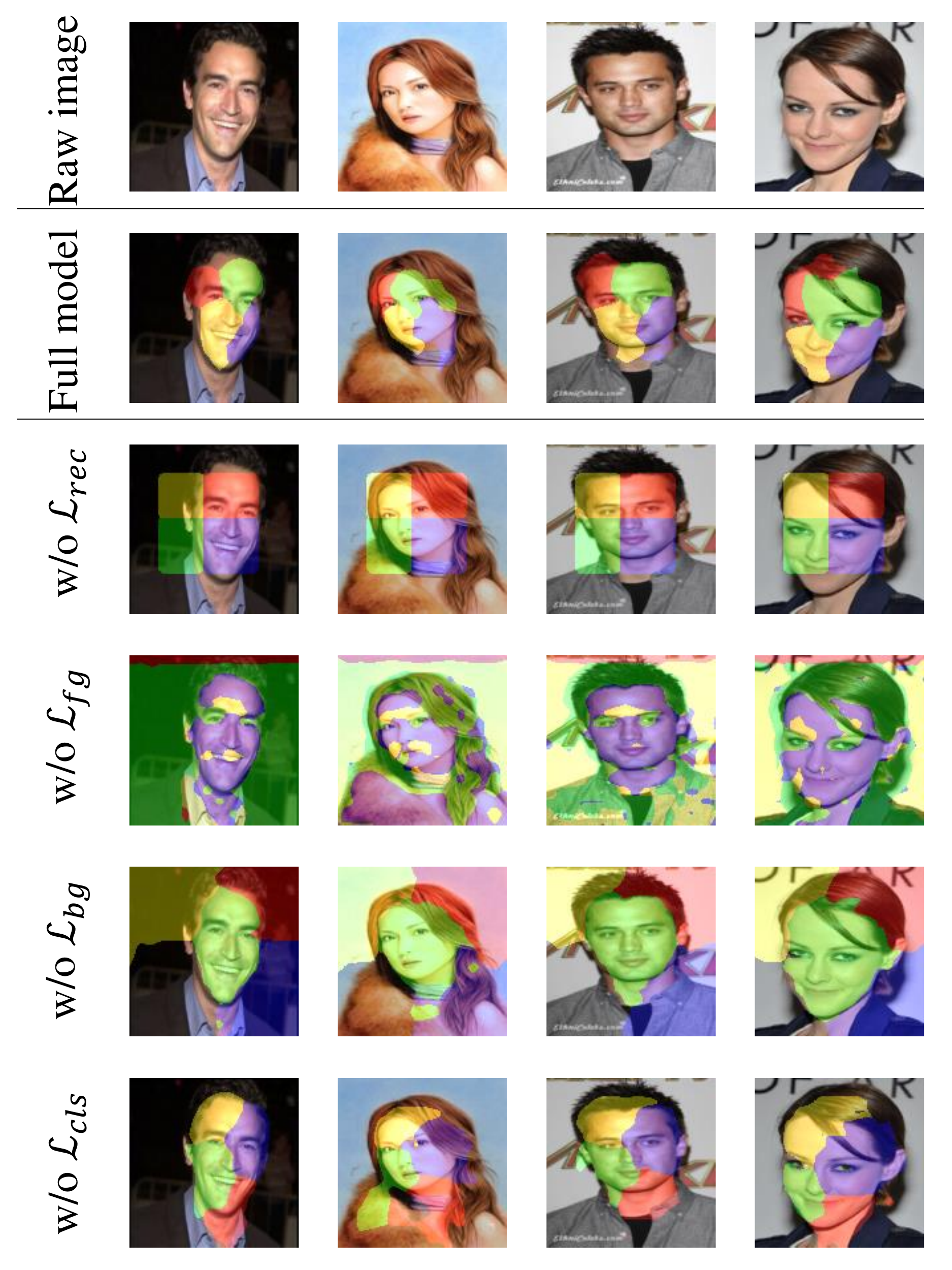}
\end{center}
\vspace{-3ex}
  \caption{Visualization and ablation studies on wild CelebA. It shows that all the $\mathcal{L}_{rec}$, $\mathcal{L}_{fg}$ and $\mathcal{L}_{bg}$ are essential for avoiding a degenerated solution and the classification loss further improves the distinctiveness of parts.}
\label{fig:celeba}
\vspace{-10px}
\end{figure}

The result shows that our method, without using any additional information, outperforms all other methods, including SCOPS which uses an additional saliency map. 
Although this is a proxy evaluation metric, it in some ways illustrates the good semantic consistency of our approach. To visualize the part segmentation result, we show some resulting images from CelebA in Fig. \ref{fig:celeba}.

\begin{table}
\vspace{-1ex}
\begin{center}
\begin{tabular}{l|cc}
\hline
Method & K=4 & K=8 \\
\hline
ULD~\cite{thewlis2017unsupervised_iccv}  & - & 31.30\\
Zhang \etal~\cite{zhang2018unsupervised}  & - & 40.82\\
IMM~\cite{jakab2018unsupervised} & 19.42 & \textbf{8.74} \\
Lorenz \etal~\cite{lorenz2019unsupervised} & 15.49 & 11.41 \\
\hline
SCOPS~\cite{hung:CVPR:2019}(w/o saliency) & 46.62 & 22.11 \\
SCOPS~\cite{hung:CVPR:2019}(with saliency) & 21.76 & 15.01 \\
Ours & \textbf{15.39} & 12.26 \\
\hline
\end{tabular}
\end{center}
\vspace{-3ex}
\caption{Landmark regression results on wild CelebA. We report the landmark regression error in terms of mean L2 distance normalized by inter-ocular distance. 
Note that the first three methods are specially designed for landmarks discovery and Lorenzet \etal is not for precise part segmentation for its elliptical shape assumption. We list them for reference.
}
\label{tb:celeba}
\end{table}

In addition to CelebA, we also test our method on another human face dataset: AFLW~\cite{koestinger11a}. AFLW contains 25K faces, each of which is annotated with 21 landmarks. AFLW is a more challenging dataset because the variation across pictures is even larger. Some other works~\cite{Thewlis2019a} provide a subset of AFLW and select five face annotations for each face. We uses their \textit{unaligned} dataset for testing. We compare our method with relevant works in Table \ref{tb:AFLW}.
In the setting of $K=8$, we achieve a mean error of $13.13$, whereas SCOPS achieves $16.05$ even with a saliency prior. 
The comparisons on two different datasets demonstrate the superiority of our proposed method in terms of semantic consistency of object parts.

\begin{table}
\vspace{-2ex}
\begin{center}
\begin{tabular}{l|c}
\hline
Method & Mean Error \\
\hline
IMM~\cite{jakab2018unsupervised} & 13.31 \\
Lorenz\etal~\cite{lorenz2019unsupervised} & 13.60 \\
SCOPS~\cite{hung:CVPR:2019}(w/o saliency) & 56.08 \\
SCOPS~\cite{hung:CVPR:2019}(with saliency) & 16.05 \\
\hline
Ours & \textbf{13.13} \\
\hline
\end{tabular}
\end{center}
\vspace{-3ex}
\caption{Landmark regression results on unaligned AFLW. All methods are in the setting of $K=8$.}
\label{tb:AFLW}
\vspace{-12px}
\end{table}

\textbf{Ablation Study.} 
\label{Sec:ablation}
To validate the effectiveness of each loss function, we conduct ablation studies on CelebA. Table \ref{tb:ablation} shows the contribution of each loss function by excluding one loss function each time. For a more intuitive understanding, the visual results of each ablation study are shown in the Fig. \ref{fig:celeba}. The results show that the combination of the reconstruction loss, the foreground loss, and the background loss can effectively help avoid a degenerated result. When any constraint is missing, the regression mean error increases dramatically. The visualization in Fig. \ref{fig:celeba} can show the impact of each loss more intuitively. The reconstruction loss $\mathcal{L}_{rec}$ is the key to close the loop for unsupervised learning and ensure semantic consistency. Without $\mathcal{L}_{rec}$, the segmentation results are no better than uniform squares. Without the foreground loss $\mathcal{L}_{fg}$, the parts learned are scattered all over an image. If we remove the background loss $\mathcal{L}_{bg}$, the result leads to another trivial solution, with one part covering the whole object and others covering the background. The classification loss $\mathcal{L}_{cls}$ helps further improve the distinctiveness of the learned part features and enforce cross-instance semantic consistency. After adding $\mathcal{L}_{cls}$, the mean error is further reduced from $14.38$ to $12.26$ as in Table \ref{tb:ablation}. Without  $\mathcal{L}_{cls}$, the part learned may cover some semantically unrelated regions. For example, the facial area contains some pixels from neck and clothes. 

As we discussed in Sec. \ref{sec:method}, the whole framework is based on an encoder-decoder structure, with the reconstruction loss to setup unsupervised learning tasks. The reconstruction loss helps ensure both the \textit{semantic constraint} and the  \textit{geometry constraint}. The squeeze-and-expand bottleneck block makes it possible to further add the classification loss for improving the semantic consistency and add the foreground and  background losses for satisfying the geometry constraint. The ablation studies demonstrate that all the four losses are essential and effective for good segmentation results.

\begin{table}
\vspace{-2ex}
\begin{center}
\begin{tabular}{l|cc}
\hline
Method & K=4 & K=8  \\
\hline
w/o $\mathcal{L}_{rec}$ & 49.32 & 46.01 \\
w/o $\mathcal{L}_{fg}$ & 45.77 & 41.42 \\
w/o $\mathcal{L}_{bg}$ & 31.83 & 29.93 \\
w/o $\mathcal{L}_{cls}$ & 17.20 & 14.38 \\
\hline
Full Model & \textbf{15.39} & \textbf{12.26} \\
\hline
\end{tabular}
\end{center}
\vspace{-3ex}
\caption{Ablation study on CelebA. The results shows that all the four losses are essential for semantically consistent part segmentation.} \label{tb:ablation}
\vspace{-2ex}
\end{table}

\begin{figure}[t]
\vspace{-1ex}
\centering

\subfloat[CUB]{
    \label{fig:CUB}
    \begin{minipage}[t]{0.45\textwidth}
        \centering
        \includegraphics[width=0.9\linewidth]{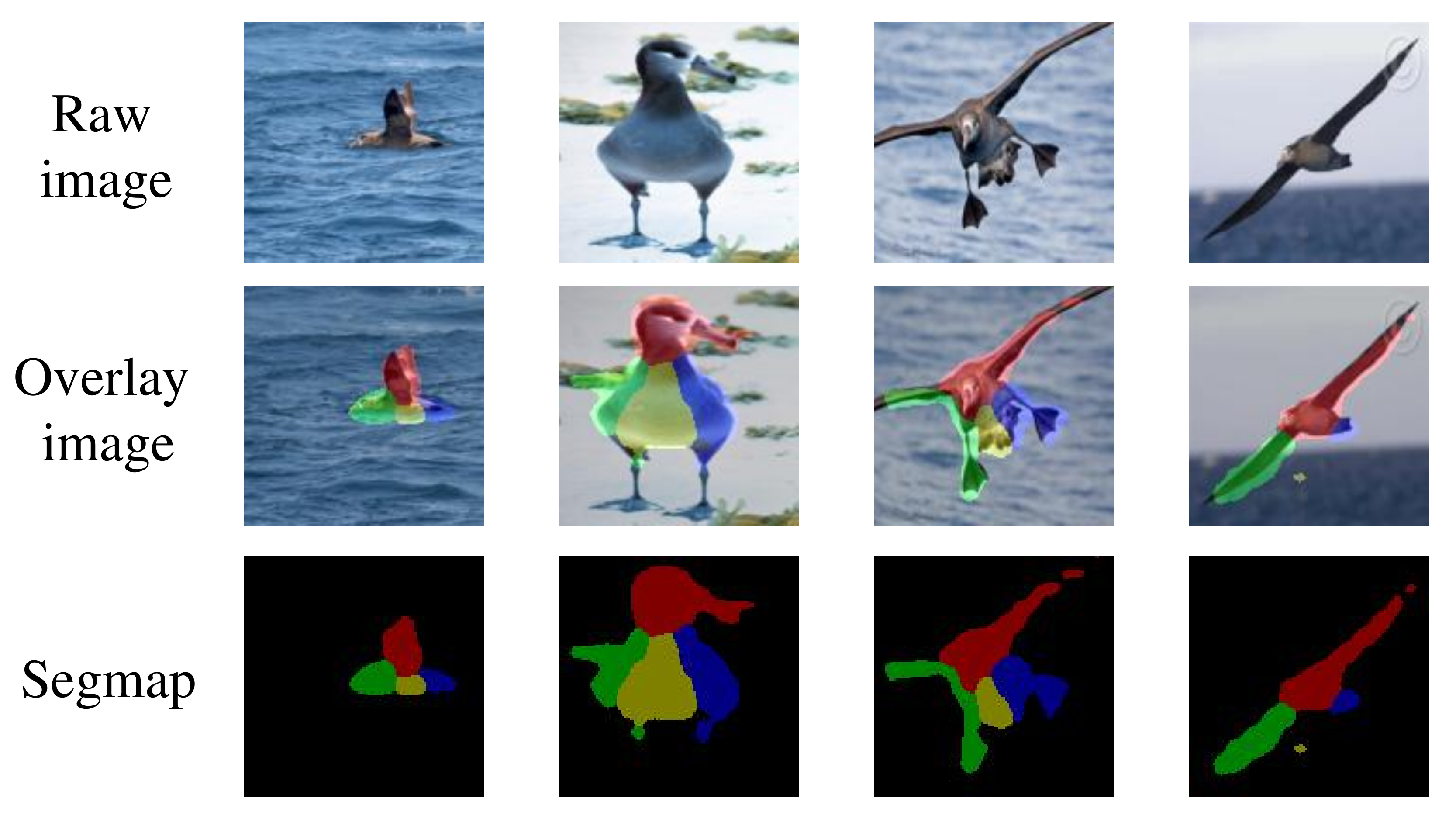}
    \end{minipage}
    }
\vspace{-2ex}
\subfloat[PASCAL VOC]{
    \label{fig:VOC}
    \begin{minipage}[t]{0.45\textwidth}
        \centering
        \includegraphics[width=0.9\linewidth]{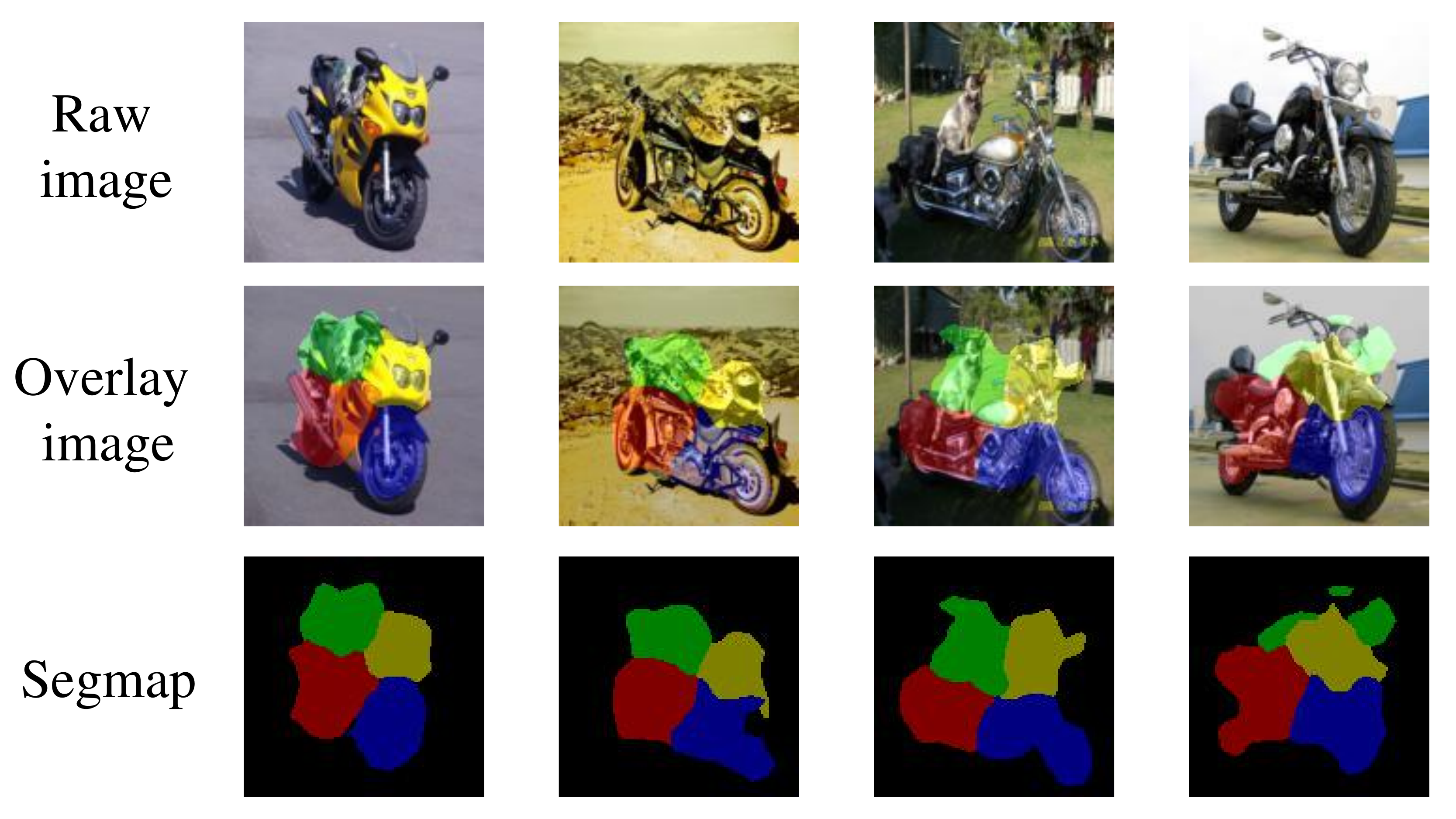}
    \end{minipage}
    }
\vspace{-1ex}
  \caption{Visualization on CUB and PASCAL VOC. Each instance (shown in one column) has three images: the input image in the first row, the segmentation map in the last row, and the overlay of segmentation result on the input image in the second row.}
\label{fig:CUB_VOC}
\vspace{-12px}
\end{figure}

\subsection{Evaluation on Common Objects}
\label{sec:exp_cub_voc}
To test the generalization ability of our method, we also evaluate it on other two datasets: CUB-200-2011~\cite{everingham2010the}, termed as CUB for short, and PASCAL VOC ~\cite{wah2011the}, termed as VOC in the following paragraphs.

CUB~\cite{everingham2010the} is a dataset with $11,788$ photos of $200$ bird species in total.
This dataset is challenging due to the large variation in bird attitudes. 
In this experiment, we follow ~\cite{hung:CVPR:2019} and select the photos from the first $3$ categories for segmentation. Similar to the previous experiments for human faces, to quantify the performance of our result, we use the expected coordinates of each part as a key point position estimation for landmark regression, where we use the annotated $15$ part locations as the ground truth landmarks. 
We show the mean error of landmark regression in Table \ref{tb:CUB}, where CUB-1, CUB-2 and CUB-3 represents the results of three categories of birds. Similarly, our results are superior to other methods, which indicates that our method keeps a good semantic consistency for objects with large variations. Visualization of CUB is shown in Fig. \ref{fig:CUB}.

Next, we test our method on PASCAL VOC~\cite{wah2011the} for more object categories. VOC is a large-scale dataset which has been widely used for object detection and semantic segmentation tasks. It contains more than $10,000$ images of $20$ categories in total, and about $7,000$ object instances annotated with segmentation maps. We follow ~\cite{hung:CVPR:2019} and select seven categories to test our method. As VOC does not have landmark annotations, we use the ground truth segmentation masks of object instances and report in Table \ref{tb:PASCAL} the $\operatorname{IOU}$ of our part-aggregated segmentation result compared with the ground truth mask. The part-aggregated segmentation is simply taken as the union of all the learned foreground parts. It is worth noting that, as SCOPS~\cite{hung:CVPR:2019} explicitly uses an additional saliency map (or silhouette), this is not a fair comparison to some extends. Nevertheless, our results still outperforms previous methods on most categories, and are better than DFF~\cite{collins2018deep} on all categories. Visualization of VOC is shown in Fig. \ref{fig:VOC}.

\textbf{Limitations}: {Note that while our method performs well in terms of image transformation and large appearance variance, we find it still struggles with distinguishing between orientations of near-symmetric objects. Distinguishing flipped object parts remains a challenging task for self-supervised learning and deserves dedicated studies in our future work.}

\begin{table}
\vspace{-2ex}
\begin{center}
\begin{tabular}{l|ccc}
\hline
Method & CUB-1 & CUB-2 & CUB-3  \\
\hline
zhang \etal~\cite{zhang2018unsupervised} & 30.12 & 29.36 & 28.19 \\
DFF ~\cite{collins2018deep} & 22.42 & 21.62 & 21.98  \\
SCOPS ~\cite{hung:CVPR:2019} & 18.50 & 18.82 & 21.07 \\
\hline
Ours & \textbf{18.15} & \textbf{17.54} & \textbf{19.40} \\
\hline
\end{tabular}
\end{center}
\vspace{-3ex}
\caption{Landmark regression results on CUB. We report the mean error of landmark regression,
where the horizontal and vertical coordinates are normalized by the width and height of ground truth bounding box, respectively.
}
\label{tb:CUB}
\vspace{-2ex}
\end{table}

\begin{table}
\begin{center}
\resizebox{82mm}{7.4mm}{
\begin{tabular}{l|ccccccc}
\hline
category & sheep & bus & motor & horse & cow & areo & car \\
\hline
DFF~\cite{collins2018deep} & 51.03  & 58.63 & 54.80 & 49.51 & 56.39 & 48.38 & 56.48 \\
SCOPS~\cite{hung:CVPR:2019} & 56.95 & 73.82 & 58.53 & \textbf{55.76} & 60.79 & \textbf{69.02} & 65.18\\
\hline
Ours &  \textbf{64.87} & \textbf{78.17} & \textbf{59.98} & 53.65 & \textbf{65.60} & 52.82 & \textbf{70.49}\\
\hline
\end{tabular}
}
\end{center}
\vspace{-3ex}
\caption{Evaluation results on PASCAL VOC. We report $\operatorname{IOU}$
of part-aggregated segmentation compared with the ground truth mask.
}
\label{tb:PASCAL}
\vspace{-2ex}
\end{table}

\subsection{Disentanglement on DeepFashion}

\label{sec:exp_fashion}
To illustrate the effect of disentanglement, which is a key component in our framework, we evaluation our method on DeepFashion~\cite{liuLQWTcvpr16DeepFashion} dataset, which contains over $800,000$ diverse images.
In our experiment, we select full-body images of their ``in-shop clothes retrieval benchmark'' for learning. 

\begin{figure}[t]
\vspace{-3ex}
\begin{center}
  \includegraphics[width=0.9\linewidth]{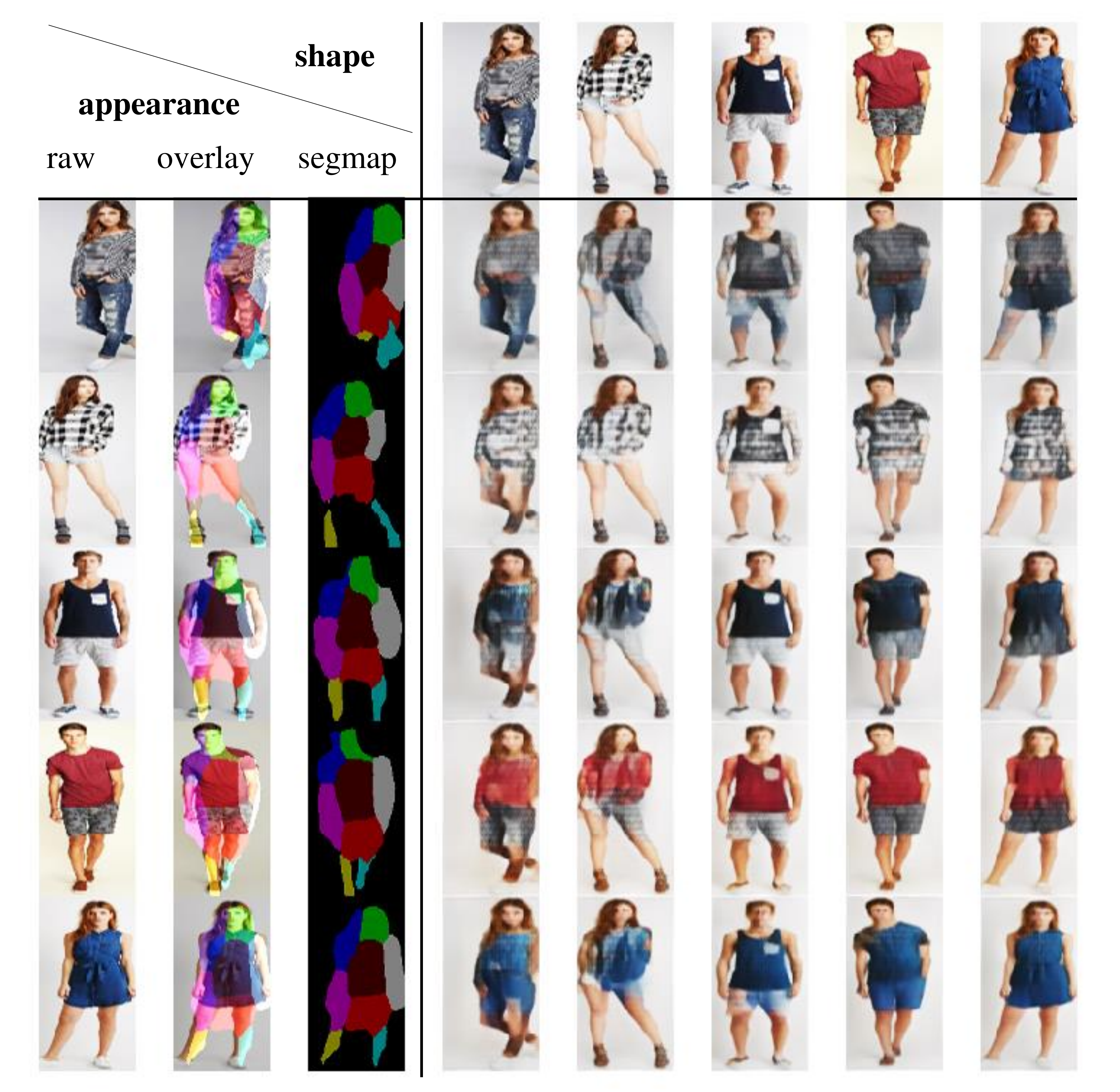}
\end{center}
\vspace{-4ex}
  \caption{Appearance and shape disentanglement on DeepFashion. Images in the same column have the same posture, but their appearances come from the leftmost images.}
\label{fig:fashion}
\vspace{-2ex}
\end{figure}

We visualize the part segmentation results and appearance transfer results in Fig. \ref{fig:fashion}. Every generated image is synthesized by the appearance taking from one image in the leftmost column and the shape taking from one image in the topmost row. The visualization is mainly to show if the appearance and shape representations are well disentangled, rather than showing high-quality synthesis images.  

\section{Conclusion}
In this paper, we have presented an unsupervised method that can learn part segmentation without supervised annotations by disentangling appearance and shape. We considered two essential constraints: \textit{semantic constraint} and \textit{geometry constraint} for a good part segmentation result, and accordingly developed an encoder-decoder neural network with a squeeze-and-expand bottleneck in the middle for disentangling the appearance and shape appearances. This design effectively removes the dependency on a saliency map or object segmentation mask, yet leads to an improved result. Comprehensive experiments on a wide variety of objects such as human faces, birds, and PASCAL VOC objects verified the effectiveness of the proposed method in terms of both segmentation quality and disentanglement result.

\section{Acknowledgment}
This work was supported by the National Key Research and Development Program of China (No. 2017YFA0700904, No.2020AAA0104304), NSFC Projects (Nos. 61620106010, 62076147, U19A2081, U19B2034, U1811461),
Beijing Academy of Artificial Intelligence (BAAI), Tsinghua-Huawei Joint Research Program, a
grant from Tsinghua Institute for Guo Qiang, Tiangong Institute for Intelligent Computing, and the
NVIDIA NVAIL Program with GPU/DGX Acceleration.

\clearpage

{\small
\bibliographystyle{ieee_fullname}
\bibliography{egbib}
}

\end{document}


\title{Appendix for: Unsupervised Part Segmentation through Disentangling\\  Appearance and Shape}
\maketitle

\appendix

\section{Implementation Details}
\subsection{TPS Transformation}
We adopt the implementation open-sourced by ~\cite{jakab2018unsupervised} for TPS transformation. The source control points are from a $10\times 10$ regular grid. All parameters are sampled from a gaussian distribution with zero mean and a given standard deviation. Each control point is perturbed by a parameter with standard deviation $0.001$, and then applied with an additional perturbation with a standard deviation $0.005$ with $50\%$ probability. For the affine component, we set the standard deviation as $0.1$ for the translation parameter and $0$ for both the rotation and scale parameters. Note that all the coordinates are normalized to be in the range $[-1, 1]$ during deformation.

\subsection{Perceptual Loss}
We use the perceptual loss as in previous works~\cite{jakab2018unsupervised, xu2020unsupervised}. An ImageNet-pretrianed VGG-19~\cite{simonyan2014very} is adopted as a feature extractor in our experiments. Given a pair of images (an input image and a reconstructed image in this work), we extract the output features of $\operatorname{input}$, $\operatorname{conv1\_2}$, $\operatorname{conv2\_2}$, $\operatorname{conv3\_2}$, $\operatorname{conv4\_2}$, $\operatorname{conv5\_2}$ layers from the pretrained VGG-19 for each image, and weighted average the $\operatorname{L2}$ distance of each pair of features. To balance the contribution of each layer, the weight of each layer is set as the reciprocal of its average $\operatorname{L2}$ loss every 100 steps, as used in ~\cite{chen2017photographic, jakab2018unsupervised}.

\subsection{Image Generation}
After the \textit{expand} operation, the $L$ dimensional appearance features belonging to the same part are almost the same, which makes reconstruction very challenging. The encoding method proposed in ~\cite{lorenz2019unsupervised} suggests that the coordinates of each pixel can help on reconstruction. Therefore, 
we normalize the coordinates of each pixel with respect to image width and height and stack all the coordinates to form a $2\times H\times W$ tensor. Then we concatenate them with the rendered appearance feature maps $\mathbf{A}^{i\to j}\in \mathbb{R}^{L\times H\times W}$ to form a new tensor $\Tilde{\mathbf{A}}^{i\to j}\in \mathbb{R}^{(L+2)\times H\times W}$. In our implementation, we use $\Tilde{\mathbf{A}}^{i\to j}$ as input to the decoder $D$ instead of $\mathbf{A}^{i\to j}$.

\subsection{Final Loss Function.} 
In the proposed method, the final loss function is a linear combination of four loss functions, including the reconstruction loss $\mathcal{L}_{rec}$, the part classification loss $\mathcal{L}_{cls}$, the foreground loss $\mathcal{L}_{fg}$, and the background loss $\mathcal{L}_{bg}$, as 

\begin{equation}
\label{eq:overall_loss_appendix}
\mathcal{L}_{sum} = \lambda_{rec}\mathcal{L}_{rec} + \lambda_{cls}\mathcal{L}_{cls} +  
 \lambda_{fg}\mathcal{L}_{fg} + \lambda_{bg}\mathcal{L}_{bg}.
\end{equation}

We use different settings for different tasks, as shown in Table \ref{tb:settings}. These combination weights were empirically obtained by coarse grid search. The models were trained on one GeForce RTX 2080 Ti GPU with 32 images per step for 30 epochs. Our model is light and efficient. On average, each model completes training in three hours.

\begin{table}
\begin{center}
\begin{tabular}{l|c|cccc}
\hline
Dataset & $K$(\# of parts) & $\lambda_{rec}$ & $\lambda_{cls}$ & $\lambda_{fg}$ & $\lambda_{bg}$ \\
\hline
CelebA  & 4/8 & 1.5 & 1.5 & 0.5 & 1.0\\
AFLW    & 4/8 & 1.5 & 1.5 & 0.5 & 1.0 \\
\hline
CUB-1/2/3 & 4 & 1.5 & 1.5 & 0.3 & 1.0 \\
\hline
VOC-car & 4 & 1.5 & 1.0 & 0.3 & 0.1 \\
VOC-bus & 4 & 1.5 & 1.0 & 0.5 & 0.1 \\
VOC-horse & 4 & 1.5 & 3.0 & 0.3 & 0.1 \\
VOC-aero & 4 & 1.5 & 3.0 & 0.3 & 0.1 \\
VOC-motor & 4 & 1.0 & 1.0 & 0.3 & 0.1 \\
VOC-cow & 4 & 1.5 & 2.0 & 0.3 & 0.1 \\
VOC-sheep & 4 & 1.5 & 2.0 & 0.5 & 0.1 \\
\hline
DeepFashion & 9 & 1.5 & 3.0 & 0.5 & 0.1 \\
\hline
\end{tabular}
\end{center}
\caption{Settings for different experiments: the number of parts and combination weights in the final loss function.}
\label{tb:settings}
\end{table}

\section{More Visualization Results}
\begin{figure*}
\begin{center}
\includegraphics[width=1.0\linewidth]{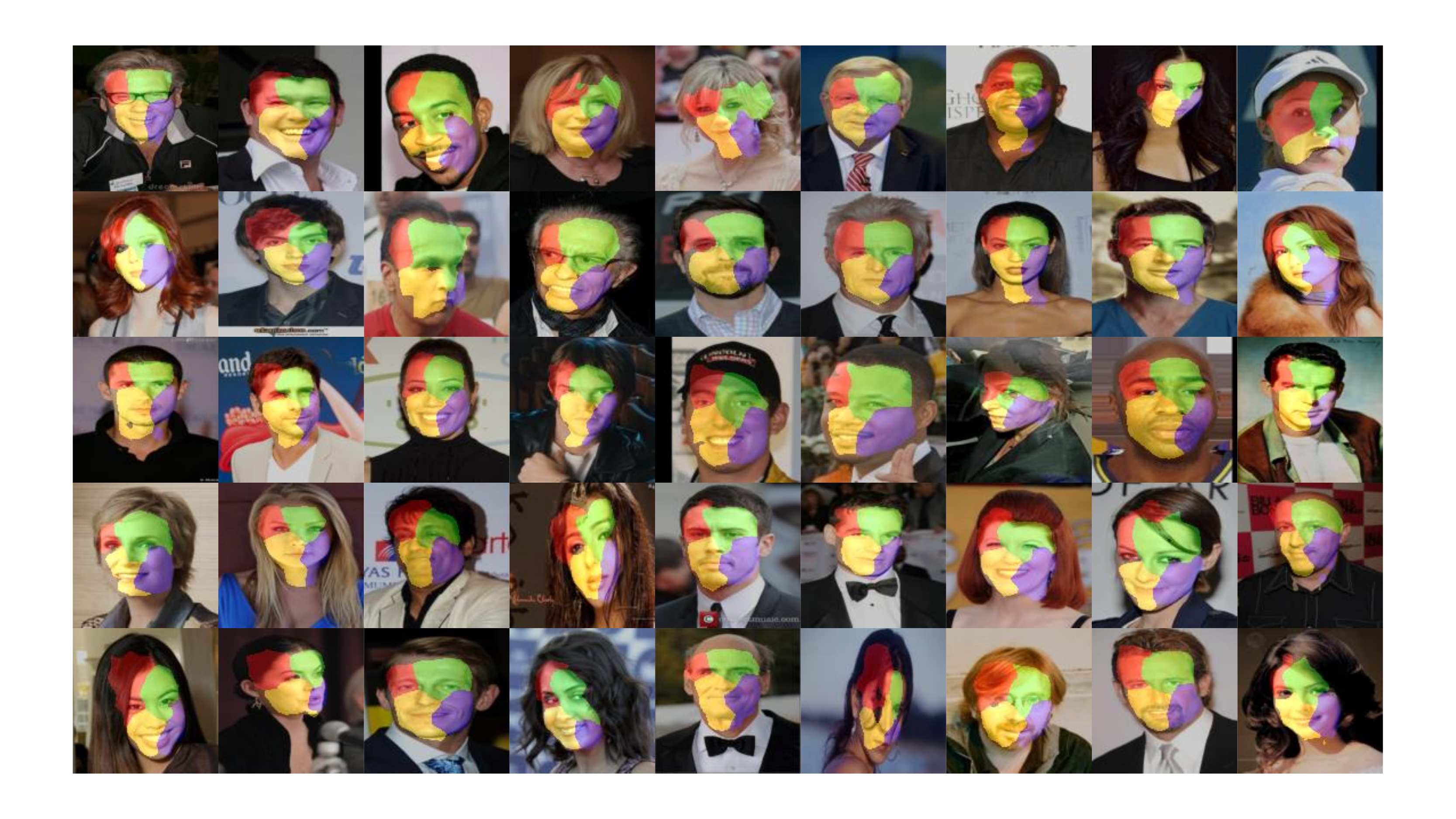}
\end{center}
  \caption{Visualization results on the wild CelebA dataset with $K=4$.}
\label{fig:appendix-celeba-4c}
\end{figure*}

\begin{figure*}
\begin{center}
\includegraphics[width=1.0\linewidth]{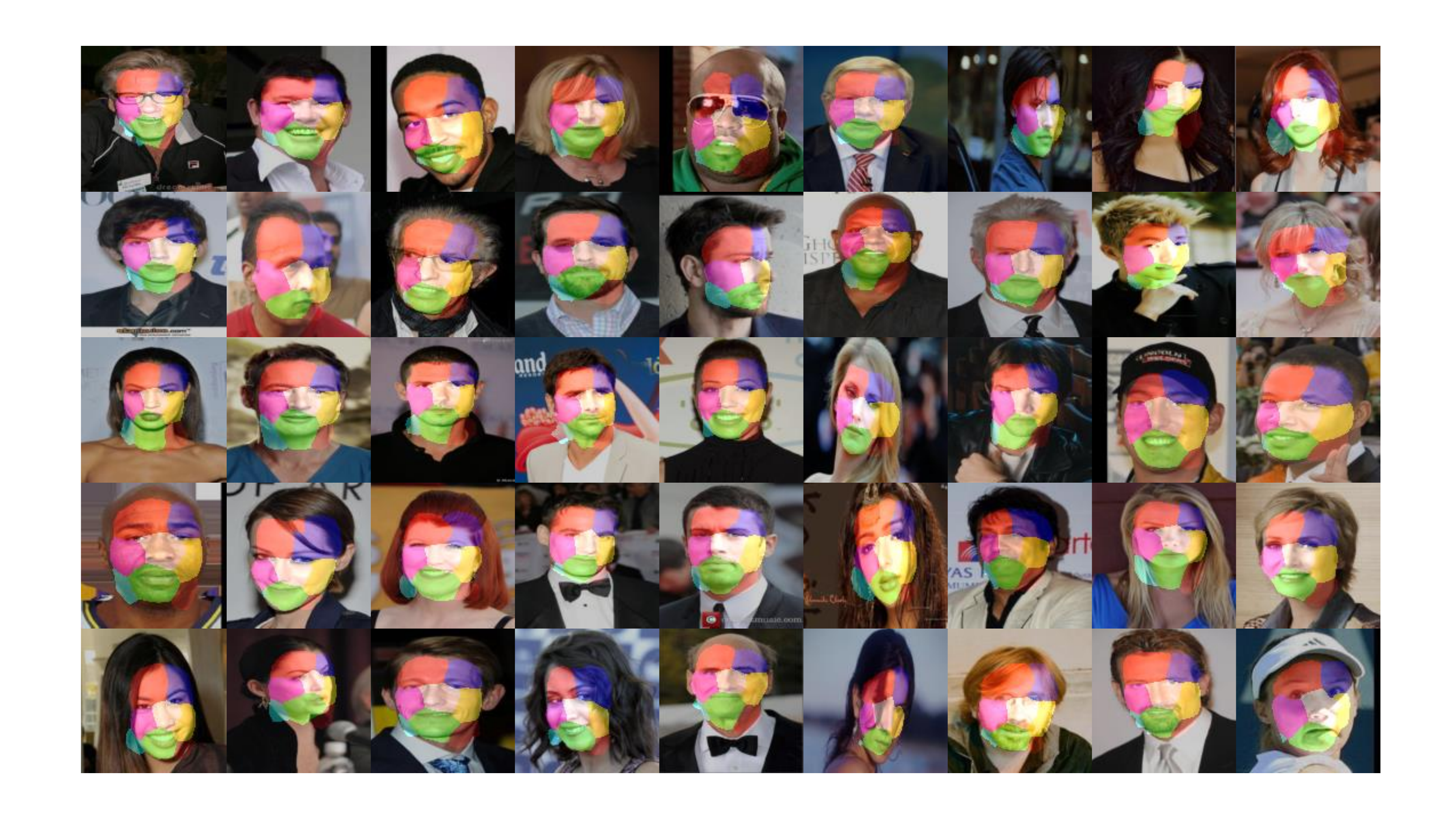}
\end{center}
  \caption{Visualization results on the wild CelebA dataset with $K=8$.}
\label{fig:appendix-celeba-9c}
\end{figure*}

\begin{figure*}
\begin{center}
\includegraphics[width=1.0\linewidth]{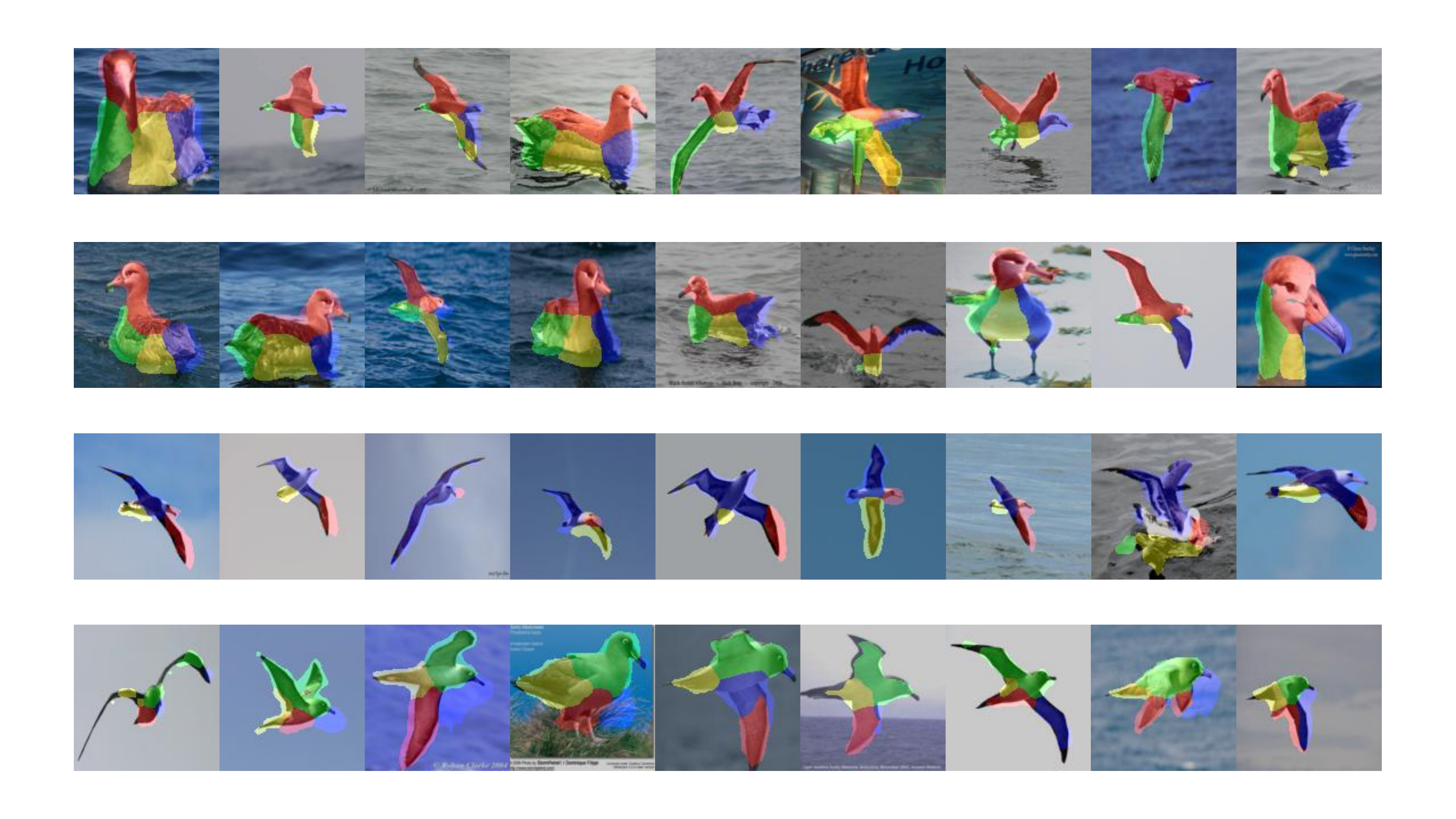}
\end{center}
  \caption{Visualization results on the CUB dataset with $K=4$.}
\label{fig:appendix-CUB}
\end{figure*}

\begin{figure*}[t]
\centering

\subfloat[Motor]{
    \label{fig:appendix-VOC-motor}
    \begin{minipage}[t]{1.0\textwidth}
        \centering
        \includegraphics[width=1.0\linewidth]{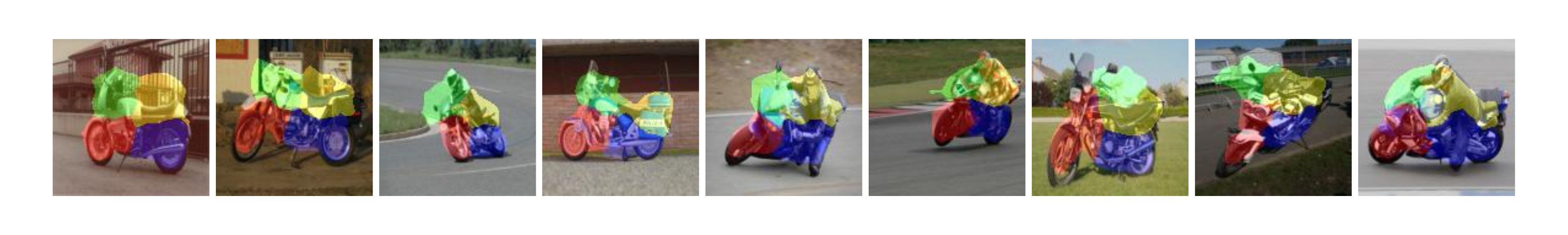}
    \end{minipage}
    }
    
\vspace{-2ex}
\subfloat[Sheep]{
    \label{fig:appendix-VOC-sheep}
    \begin{minipage}[t]{1.0\textwidth}
        \centering
        \includegraphics[width=1.0\linewidth]{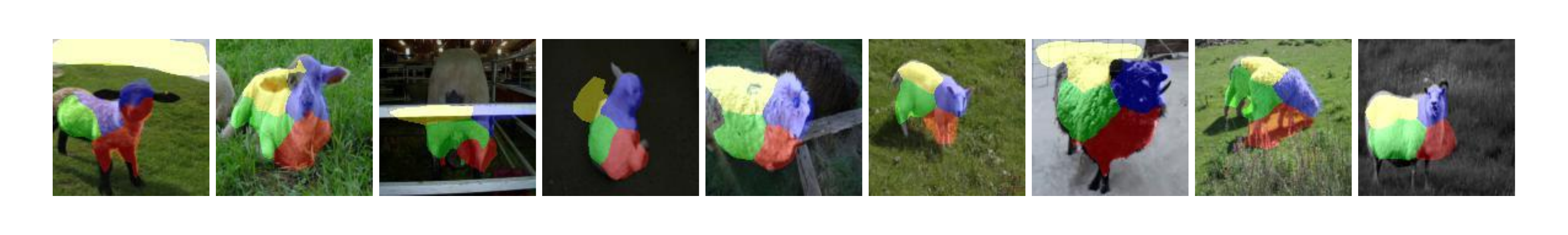}
        
    \end{minipage}
    }

\vspace{-2ex}
\subfloat[Bus]{
    \label{fig:appendix-VOC-bus}
    \begin{minipage}[t]{1.0\textwidth}
        \centering
        \includegraphics[width=1.0\linewidth]{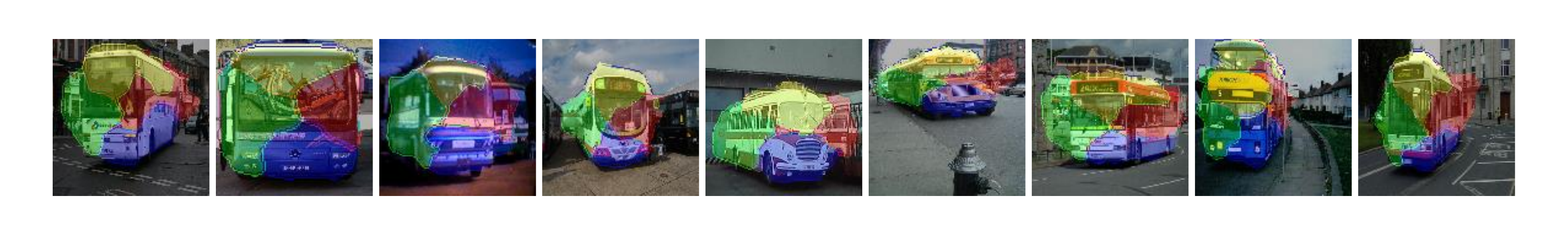}
    \end{minipage}
    }

\vspace{-2ex}
\subfloat[Aero]{
    \label{fig:appendix-VOC-aero}
    \begin{minipage}[t]{1.0\textwidth}
        \centering
        \includegraphics[width=1.0\linewidth]{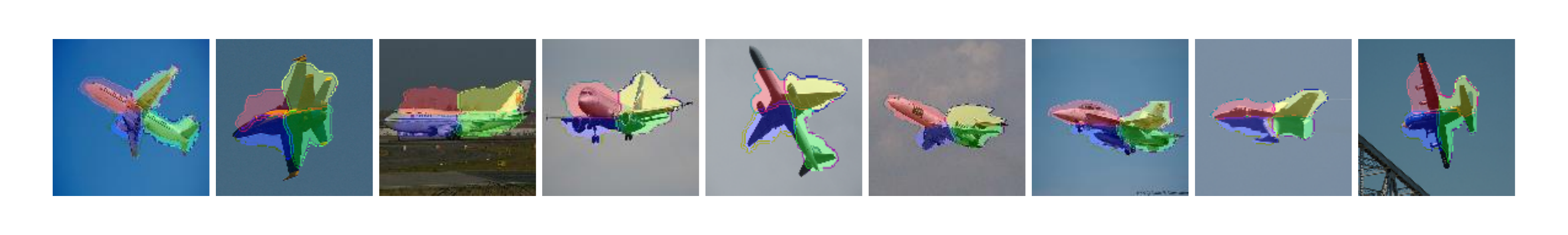}
    \end{minipage}
    }

\vspace{-2ex}
\subfloat[Cow]{
    \label{fig:appendix-VOC-cow}
    \begin{minipage}[t]{1.0\textwidth}
        \centering
        \includegraphics[width=1.0\linewidth]{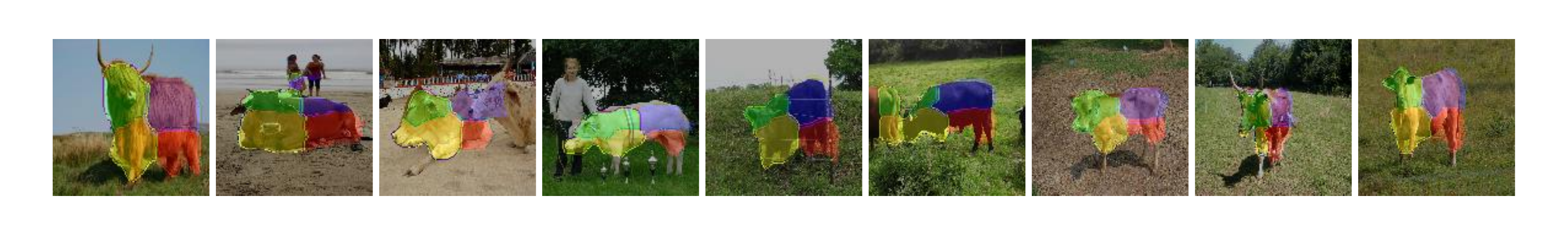}
    \end{minipage}
    }
    
\vspace{-2ex}
\subfloat[Horse]{
    \label{fig:appendix-VOC-horse}
    \begin{minipage}[t]{1.0\textwidth}
        \centering
        \includegraphics[width=1.0\linewidth]{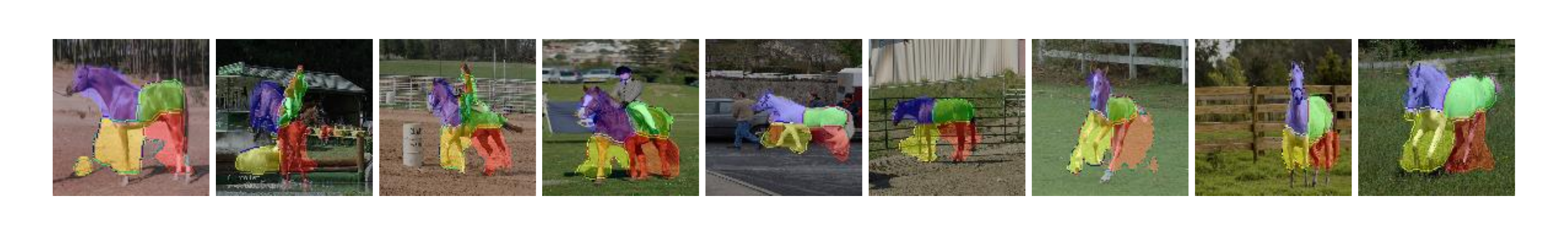}
    \end{minipage}
    }

\vspace{-2ex}
\subfloat[Car]{
    \label{fig:appendix-VOC-car}
    \begin{minipage}[t]{1.0\textwidth}
        \centering
        \includegraphics[width=1.0\linewidth]{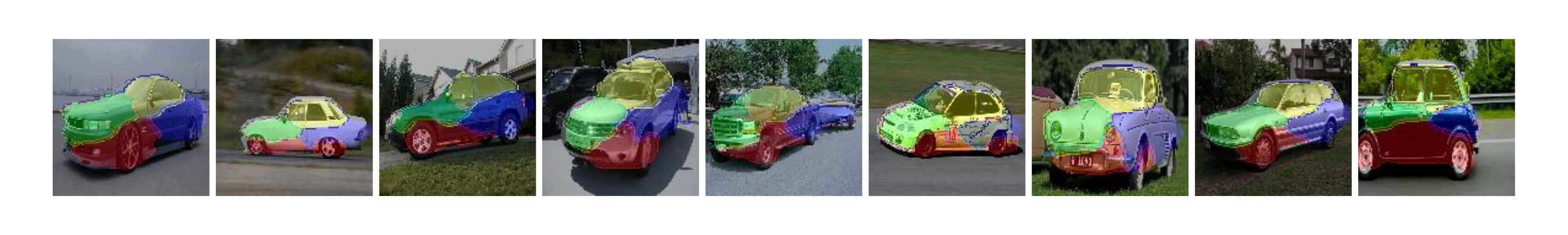}
    \end{minipage}
    }

  \caption{Visualization results on PASCAL VOC.}
\label{fig:appendix-VOC}
\end{figure*}

We provide more visualization results on wild CelebA~\cite{liu2015faceattributes}, CUB~\cite{everingham2010the}, and PASCAL VOC~\cite{wah2011the} from Fig. \ref{fig:appendix-celeba-4c} to Fig. \ref{fig:appendix-VOC}. These results indicate that our method keeps a good semantic consistency for objects with large variations.

\clearpage

\clearpage

\clearpage

\clearpage

{\small
\bibliographystyle{ieee_fullname}
\bibliography{egbib}
}